\RequirePackage{fix-cm}
\documentclass[smallextended]{svjour3}       
\smartqed  
\usepackage{graphicx}
\usepackage{subcaption}
\usepackage{tabularx}
\usepackage{natbib}
\usepackage{booktabs}
\usepackage{microtype}
\usepackage{multirow}
\usepackage{enumitem}
\usepackage[utf8]{inputenc}
\usepackage[T1]{fontenc}
\usepackage{soul}
\usepackage{authblk}
\usepackage{float}
\usepackage{amssymb}
\usepackage{tabu}
\usepackage{url}
\usepackage[dvipsnames]{xcolor}
\usepackage[toc,page]{appendix}
\usepackage{enumitem}
\usepackage{bbding}

\begin{document}

\title{Between welcome culture and border fence
\thanks{We acknowledge funding by the Deutsche Forschungsgemeinschaft (DFG, German Research Foundation) – 375875969 through MARDY (Modeling
Argumentation Dynamics) within SPP RATIO
and by Bundesministerium für Bildung und
Forschung (BMBF) through E-DELIB (Powering
up e-deliberation: towards AI-supported moderation).}
}
%

\subtitle{A dataset on the European refugee crisis in German newspaper reports.}


\author{Nico Blokker, Andr\'e Blessing, Erenay Dayanik, Jonas Kuhn, Sebastian Pad\'o, Gabriella Lapesa}

\authorrunning{Blokker, Blessing, Dayanik, Kuhn, Pad\'o, Lapesa} 

\institute{
Nico Blokker (\Envelope)
\at 
University of Bremen, Research Center on Inequality and Social Policy \\
\email{blokker@uni-bremen.de} 
\and
Andre Blessing, Erenay Dayanik, Jonas Kuhn, Sebastian Pad\'o, Gabriella Lapesa \at
University of Stuttgart, Institute for Natural Language Processing \\
\email{gabriella.lapesa@ims.uni-stuttgart.de} 
}

\date{Received: date / Accepted: date}

\maketitle

\begin{abstract} Newspaper reports provide a 
rich source of information on the unfolding of public debate on specific policy fields
that can serve as basis for inquiry in political science. Such debates are
often triggered by critical  events, which attract public attention and incite the reactions of  political actors: crisis sparks the debate. However,
due to the challenges of reliable annotation and modeling, few large-scale datasets with high-quality annotation are available.  

This paper introduces \textit{DebateNet2.0}, which traces the political discourse on the European refugee crisis in the German quality newspaper \textit{taz} during the year 2015. The core units of our annotation are political claims (requests for specific actions to be taken within the policy field) and the actors who make them (politicians, parties, etc.). The contribution of this paper is twofold. First, we document and release \textit{DebateNet2.0}
along with its companion R package, \texttt{mardyR}, guiding the reader through the practical and conceptual issues related to the annotation of policy debates in newspapers.  
Second, we outline and apply a Discourse Network Analysis (DNA) to \textit{DebateNet2.0}, comparing two crucial moments of the policy debate on the `refugee crisis': the migration flux through the Mediterranean in April/May and the one along the Balkan route in September/October. 
Besides the released resources and the case-study, our contribution is also methodological: we talk the reader through the steps from a newspaper article to a discourse network, demonstrating that there is not just one discourse network for the German migration debate, but multiple  ones, depending on the topic of interest (political actors, policy fields, time spans).
\keywords{Discourse Network Analysis \and Policy Debates \and Annotation \and Immigration}
\end{abstract}

\section{Introduction}
\label{intro}

\paragraph{Text-as-data for Political Claims Analysis.}
In recent years, the topic of immigration has increasingly moved into the focus of the public's interest. This is a development that is not least driven by the so called European `refugee crisis' in 2015 that affected citizens at multiple levels (practically, but also emotionally) and thus kept both law makers and the media in suspense. Politicians and public figures responded to the growing numbers of refugees with rapidly changing policy proposals, which were reported and discussed in the media. Newspapers are an extremely valuable source for the empirical investigation of the effects of such crises on policy debates, because of the fine-grained representation they provide, both at the level of content (extensive, even redundant reports of the positions of politicians and parties) and at the level of time (multiple articles per day).  

Newspaper corpora are therefore one of the primary sources used by text-as-data approaches within Political Science to study the dynamics of policy debates in order to understand coalition building and decision-making processes. Typical research questions are: What are the driving actors and prominent issues of the debate? Which constellations herald turning points of the discussion? Will new coalitions emerge or old ones prevail? As these examples illustrate, the focus of this approach is generally on political actors and their positions on the specific policy issues (referred to as political claims in the literature, \citealt{koopmansPoliticalClaimsAnalysis1999}). Unfortunately, the text-as-data approach suffers from the lack of available datasets which are both large-scale and provide high-quality annotation, qualities that are needed in order to make robust generalizations.

\begin{figure}[t]
 \centering
   \includegraphics[width=0.75\linewidth]{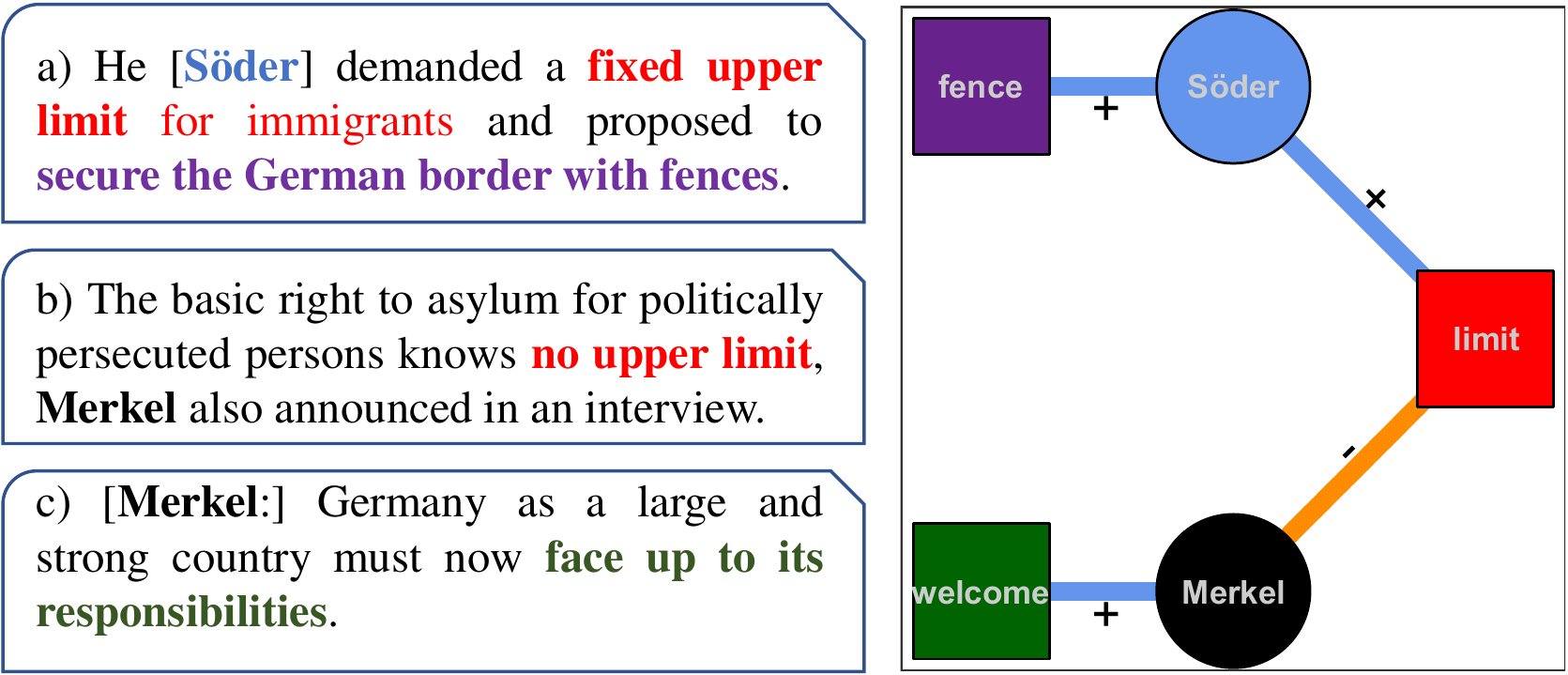}
   \caption{From text to networks: Textual spans containing political claims translated from articles in the German newspaper \textit{taz} (left panel) mapped to the corresponding network representation (right panel). Circles indicate actors and squares claims. Edges express either support ($+$, blue) or rejection ($-$, orange).}
   \label{fig:example}
\end{figure}

The work reported in this paper evolves around a dataset for this type of investigation, which we call \textit{DebateNet2.0}\footnote{An earlier version of the dataset was released in \citet{lapesa-etal-2020-debatenet} as \textit{DEbateNet-mig15}, to which we will refer as \textit{DebateNet1.0}.} and which is released with this paper. It contains high quality annotation on the topic of immigration in Germany in 2015, based on a collection of articles from a quality newspaper: \textit{die Tageszeitung (taz)}. 
 The annotation targets claims (reported demands or propositions) made by political actors (politicians, parties, organizations, etc.) regarding specific actions to be taken within the heterogeneous policy field of immigration, covering aspects of e.g.  migration control, foreign policy, integration, solidarity, racism, and others. The leftmost panel of Figure \ref{fig:example} illustrates an example of the annotation:
 the 3 textual spans contain claims attributed to two actors, Markus Söder and Angela Merkel. In the first span (a) Söder makes two claims: demanding to fix an upper limits for immigrants (red highlight) and to build a border fence (purple highlight). In the second span (b), Merkel claims the opposite regarding the upper limit, and in (c) pushes for a so-called `refugees welcome' policy strategy (green).  

Claims in \textit{DebateNet} are labelled according to an annotation schema (codebook) defined by political science experts, containing approximately 100 categories and thus providing an abstract, but nevertheless extremely fine-grained, representation of the discourse. Such a fine-grained approach to annotation supports political scientists in gaining a deeper understanding of democratic decision making in the light of discourse dynamics. 

The fine-grained annotation of actors and their claims described in the previous paragraph does not constitute an answer to the research questions about discourse dynamics we mentioned earlier, e.g., actor coalitions, turning points in the discussion. What is needed is an abstract representation of the discourse and a methodology to `mine' the pattern of the discourse. 
 We employ the Discourse Network Analysis (DNA) framework \citep{leifeldDiscourseNetworkAnalysis2016}, a text-as-data methodology which has become increasingly popular in Political Science that brings together qualitative discourse analysis (QDA) and network analysis. DNA builds on the assumption that policy debates can be modeled as a bipartite network as the one depicted in the right panel of Figure \ref{fig:example}. The network contains two types of nodes: actors (round) and claim categories (square). Edges between the nodes and the actors encode the fact that the actor made a statement regarding the specific claim categories, along with the polarity of this statement. Even if no direct edges are established between actors, or between claims, this representation allows to explore the status of specific nodes exactly by looking at its relations to the nodes of the other type. In our toy example, the upper limit claim is adequately captured as a (contested) one, as it allows us to contrast the two actors on the stage. One of the contributions of this paper, along with the release of \textit{DebateNet2.0}, is a step-wise description of the workflow that is necessary to go from textual data to discourse networks. 

Along with \textit{DebateNet2.0}, we also release its companion R package, \texttt{mardyR}\footnote{The name is an acronym for \textit{Modeling Argumentation Dynamics of Political Discourse} (in R), which is also the name of the research project within which this research was carried out (cf. acknowledgments).}:
it supports both the basic querying of the annotation (e.g., all the claims made by Angela Merkel on a certain time span) and the visualization of the discourse networks, allowing the user to trace policy debates by linking actors to their claims in textual sources, with the ultimate goal of capturing a comprehensive picture of the discourse over time. 
Crucially, we demonstrate the different insights that can be gained by parametrizing the network at different levels: focusing on actors vs. on specific claim categories vs. on actor/claims to carry out comparisons between time slices.

 In the second part of the paper, we discuss the results of our DNA case study, targeted at the characterization of the crisis/debate interplay. In particular, we focus on two stages of the refugee crisis: Spring 2015, when a large number of migrants tried to reach Europe via the Mediterranean, resulting in many lives being lost at sea; fall 2015, when the migration flux followed the Balkan route, resulting in large number of refugees at the European borders and increasingly restrictionist migration policies. Beyond the results of our analysis, this case study also serves as an illustration of the analysis of \textit{DebateNet2.0} with \texttt{mardyR} and as a guideline for the interpretation of the output of the queries of \texttt{mardyR}  on  \textit{DebateNet2.0}.     

\paragraph{Contributions.} The contributions of this paper 
comprise multiple levels. At the level of resources, we document and release \textit{DebateNet2.0} and \texttt{mardyR}. \textit{DebateNet2.0} addresses the desiderata for a large-scale high-quality policy debate dataset, namely (a) an annotation schema developed by domain experts, which are capable of capturing the fine-grained nature of the policy making process; (b) the availability of annotation layers which allow for a meaningful parametrization of the queries (e.g., for political actors, mapping of different mentions to a canonical form, and, ideally, to a party: `Die Kanzlerin', `Merkel' $\rightarrow$ Angela Merkel $\rightarrow$ CDU); (c) the availability of tools and theoretically motivated case studies which can serve as guidelines for the visualization and a meaningful interpretation of the results of the query.  

At the experimental level, we discuss the results of our case study comparing the discourse network representations of two crucial stages of the refugee crisis in Germany. 

At the methods level, we provide a step-by-step illustration of the process of deriving discourse network representations from texts and discuss the conceptual and practical issues related to this process. Additionally, we provide the reader with guidelines for the use of our dataset and for the interpretation of the output of its analysis; such guidelines can and should be considered as an additional resource we contribute together with \textit{DebateNet2.0} and \texttt{mardyR}.  

\paragraph{Contextualization.}
We believe that our contributions are highly relevant not only for researchers interested in the immigration topic which is at the focus of our investigation, but both for the NLP and the Political Science community in general. From the NLP perspective, the pipeline which leads from raw text to discourse networks is one which is full of challenges and has thus the potential to trigger progress in many sub-tasks, most notably the development of supervised classification methods able to handle fine-grained category sets in low resource scenarios and the integration of knowledge bases to support actor mapping. Besides, such a  fine-grained debate representation can serve as a blueprint to structure the understanding of how the same debate unfolds on less `well-behaved' data and and less traceable actors, as it is the case in social media.   

From the perspective of Political Science, we take a further step in deepening the understanding of highly dynamic debates on the heels of a (perceived) crisis. With the help of NLP, the dataset provided here builds the foundation to scale-up the annotation of political text both with regard to text types (genres) and scope (topic and volume).

\paragraph{Plan of the paper.}
The paper is structured as follows: Section 2 provides a background for the Discourse Network Analysis framework, along with a summary of related work within the NLP literature. Section 3 describes \textit{DebateNet2.0} in all details, from the selection of the source data, to the annotation, further processing, and quantitative analysis. Section 4 is dedicated to  \texttt{mardyR}. In section 5, we proceed to our case study: the comparison of the discourse networks from mid April to May (Mediterranean route) and late September to mid November (Balkan route). Section 6 concludes the paper by summarizing its findings and contributions and outlining further modeling work carried out on the dataset. 

\section{Background: Claims Analysis, Discourse Networks, and the Role of NLP}
\label{background}

Understanding the structure and evolution of political debates is
essential for understanding democratic decision making, and is therefore
of central interest to political science
\citep{dewildeNoPolityOld2011,zurnPoliticizationWorldPolitics2014,haunssEntstehungPolitikfeldernBedingungen2015}.

Democratic decision making can broadly follow two logics, one of which
is a `technocratic' mode, where decisions are taken by
administrative staff and field-specific experts. We focus on the
second type of decision making, the `politicized' mode, which
proceeds through programmatic statements
\citep{schmidtPolicyChangeDiscourse2004} and political debates.
While there is no general theory about mechanisms driving political discourse, there seems to be at least general agreement that the formation and evolution of discourse coalitions is a core mechanism
\citep{hajerDiscourseCoalitionsInstitutionalization1993,sabatierAdvocacyCoalitionFramework2007} and that change in these coalitions is influenced by external events and by the discourse itself \citep{leifeldDiscourseNetworkAnalysis2016}.

One promising way to gain insight 
into such discourse dynamics in 
an empirically robust fashion,
based on widely available
newspaper corpora,
combines political claims analysis
\citep{koopmansPoliticalClaimsAnalysis1999} with \textit{discourse
  network analysis} \citep{leifeldPoliticalDiscourseNetworks2012}. The unit of analysis is the \textit{claim}, that is, a demand,
proposal, or criticism that is \textit{supported} or \textit{rejected} by an \textit{actor} (a person or a group of persons) and can be \textit{categorized} with regard to its contribution to the debate at hand. Crucially, not all statements concerning the topic are to be considered a claim, but only those which target a specific action to be taken (e.g., giving empty flat to refugees). Claims and the actors who make them are represented together in a bipartite \textit{affiliation network}. A discourse coalition is
then the projection of the affiliation network on the actor side,
while the projection on the concept side yields the argumentative
clusters present in the debate.

Clearly, manual annotation of
such claims and claim-actor relations
is a resource intensive process. It it therefore natural
to ask if Natural Language Processing can help: What are
the potentials, limitations, and the practical issues of applying
NLP to the automatic construction of discourse networks?

At a general level, the NLP take on  debate modeling can build on the insights from argumentation mining and subjectivity analysis \citep{Peldszus2013,Ceron2014,swanson2015,Stab2017,Vilares2017}. An ideal NLP tool would automatically identify the actors and their contributions to the debate, and analyze such contributions at a structural level (identifying argumentative structure in their statements), at a semantic level (classifying statements into relevant categories), and at a pragmatic level (detecting the polarity of the statements).

Based on our experience \citep{pado19}, however, we argue that this task cannot be completely automated, at least not if the goal is to acquire representations at the level of (fine) granularity and quality which are required for the political science analysis. What proved successful is instead the integration of manual annotation and NLP methods \citep{blessing19} into a semi-automatic
procedure that speeds up the manual work
by providing intelligent proposals, 
efficient annotation interfaces, and
adding automatic `pre-annotation' that
is clearly labeled as such. We find that
this approach scales up with manageable loss in fine-grainedness and quality and
can serve as an example of successful
`mixed methods' that
are becoming more prominent at the intersection where 
big data meets humanities and social sciences \citep{Kuhn-LREV-to-appear}. 

\section{DebateNet2.0}
\label{data}

In the following sections we introduce the dataset and highlight the steps undertaken in its creation, from article selection over annotation specifics, to post-processing, followed by a descriptive analysis. \textit{DebateNet2.0} is available for download as a CLARIN resource (PID: \url{http://hdl.handle.net/11022/1007-0000-0007-DB07-B}).

\subsection{Source Corpus}

The source corpus consists of newspaper articles from the German quality newspaper \textit{die Tageszeitung (taz)} spanning the entire year of 2015. We choose newspaper articles for three main reasons. 

First, newspaper articles constitute an elite discourse that affects democratic decision making \citep{Schneider_Nullmeier_Hurrelmann_2007,page1996}: they aggregate claims of multiple actors while simultaneously filtering content less relevant to the discussion at hand. We are aware that this gatekeeper function \citep{white1950} of the media is a double-edged sword, as the selection and the framing of the reported news can vary significantly in relation to the editorial line of the newspaper \citep{hagen1993}, and is also manipulated at higher levels (\textit{agenda-setting}). We chose \textit{taz}, which, while left-leaning, is nevertheless considered a quality newspaper, meaning that it is reasonable to assume that claims form central actors from the entire political spectrum are reported.\footnote{We are currently extending the annotation to the more right-leaning Frankfurter Allgemeine Zeitung (FAZ) in order to compare the discourses represented in both newspapers. However, because of copyright restrictions, we will not be able to make the FAZ data publicly available.} 

Second, newspapers offer a broader overview of the discourse than most other political texts, such as party manifestos or parliamentary speeches. Hence they combine partisan, institutional and civic representatives on the level of individual actors and organizations. 

Third, this wide range of political entities enables and requires a fine-grained analysis of the issue at hand, that translates well into the use of methods such as Discourse Network Analysis \citep{leifeldDiscourseNetworkAnalysis2016}. Such breadth could be found also by gathering social media activity of politicians and MPs: the focus on policy-making would be, however, definitely less sharp and retrieving the claims would be more challenging (as the journalists both filter and highlight the relevant information).

\subsection{Article Selection}

\begin{figure}[t]
 \centering
   \includegraphics[width=.9\linewidth,height=.5\textwidth]{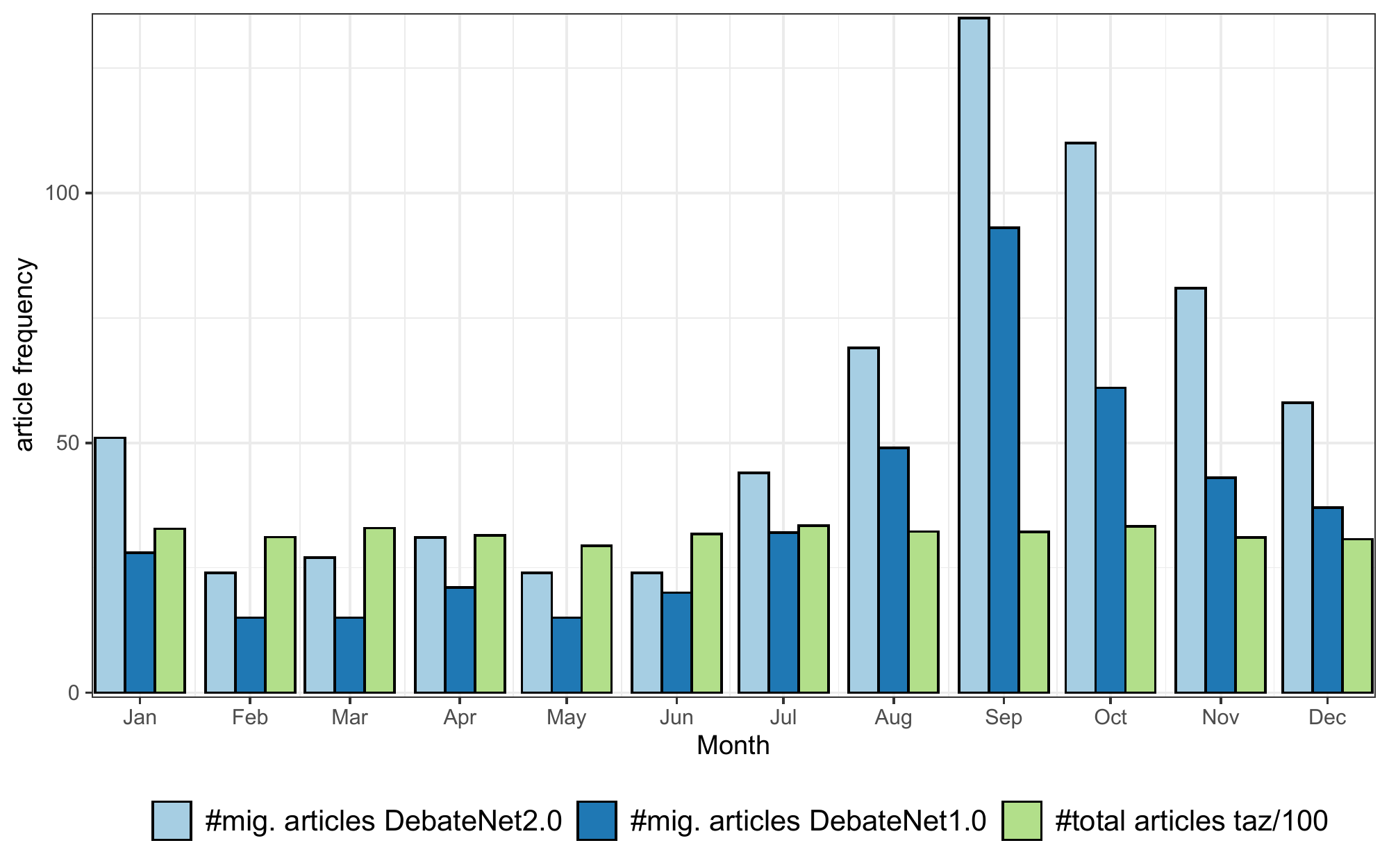}
   \caption{Number of articles on migration and total articles per month}
   \label{fig:timeline}
\end{figure}

The entire corpus contains around 38.000 articles from which roughly 700 contained claims on migration politics in Germany and were therefore selected and subsequently manually annotated (for more details, see Blessing 2019). More specifically, we defined a set of seed keywords connected to the topic (asylum, refugee, migrant/migration, etc.) and German politics. Subsequently, we trained a binary classifier \citep{joulin2016bag}, optimized on recall, on the articles found in this fashion to identify articles not covered by the keyword-based approach. 
Additionally, annotators then flagged false positive articles for being off topic. An additional criterion was applied to the subsequently presented dataset: Articles are only included when they contain at least one political claim. Figure \ref{fig:timeline} (page \pageref{fig:timeline}) illustrates two developments. Firstly, while the total number of 
articles published remains relatively stable over the months (averaging around 3.200 articles, sd = 119) the number of articles attributed to the topic of migration fluctuates considerably (mean = 37, sd = 57). Especially between August and December there is a substantial increase in migration relevant articles, peaking in September. This indicates that the newspaper placed more relative attention in this time-frame on the issue of migration at the expense of other topics. Secondly, the difference in numbers between \textit{DebateNet1.0} and \textit{DebateNet2.0} appears relatively uniformly distributed across months. Therefore, the previous release can be seen as a random sample of the complete dataset, from which similar results might carry over. Note however that this does not imply
that the content of \textit{DebateNet1.0} is also a representative sample of \textit{DebateNet2.0} (cf. Section \ref{DNA}).

Figure \ref{fig:timeline} contrasts the total number of articles the taz published in 2015 (divided by 100, right columns) with the number of articles that were deemed relevant to the migration debate in Germany and annotated accordingly (left columns). To highlight the differences in numbers between this release of \textit{DebateNet2.0} and the previous one, \textit{DebateNet1.0}, \citealt{lapesa-etal-2020-debatenet}),
the middle column represents the number of relevant articles in the earlier release.

\subsection{Annotation}

\begin{figure}[t!]
\centering
\begin{subfigure}[b]{1\textwidth}
   \includegraphics[width=1\linewidth]{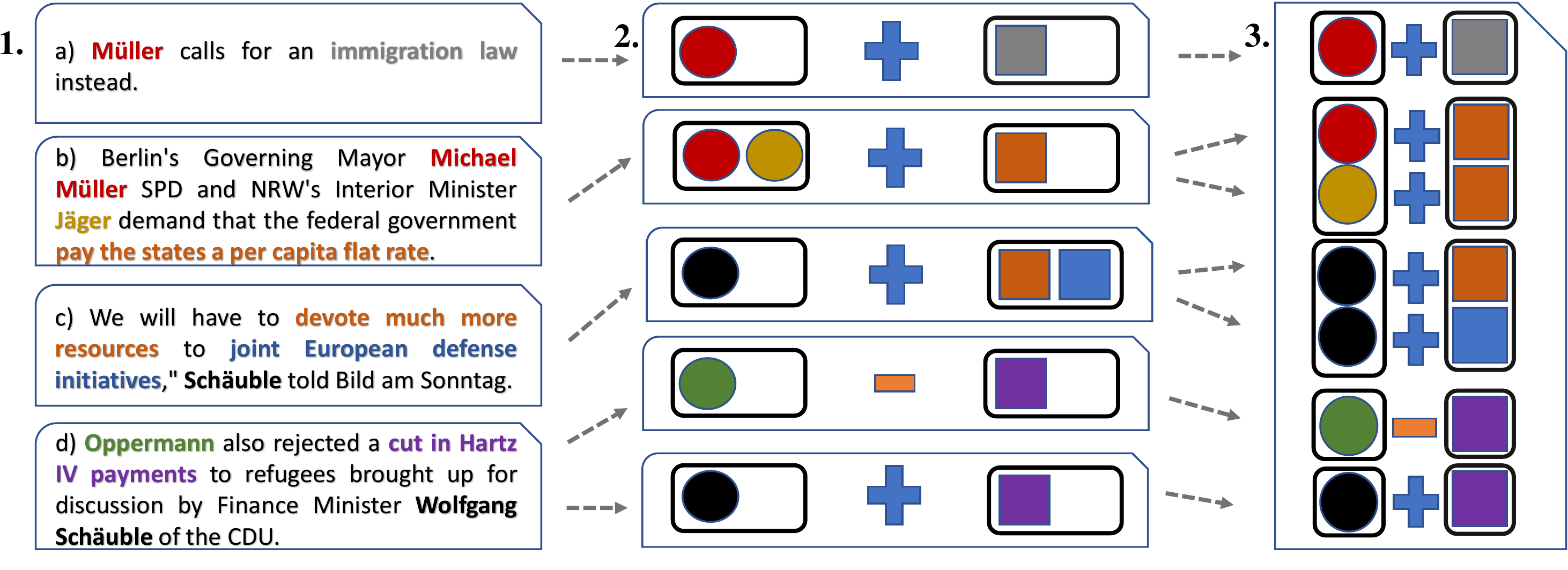}
   \caption{1. text snippets; 2. annotations, 3. observations}
   \label{fig:concepta} 
\end{subfigure}
\begin{subfigure}[b]{.5\textwidth}
   \includegraphics[width=.9\linewidth]{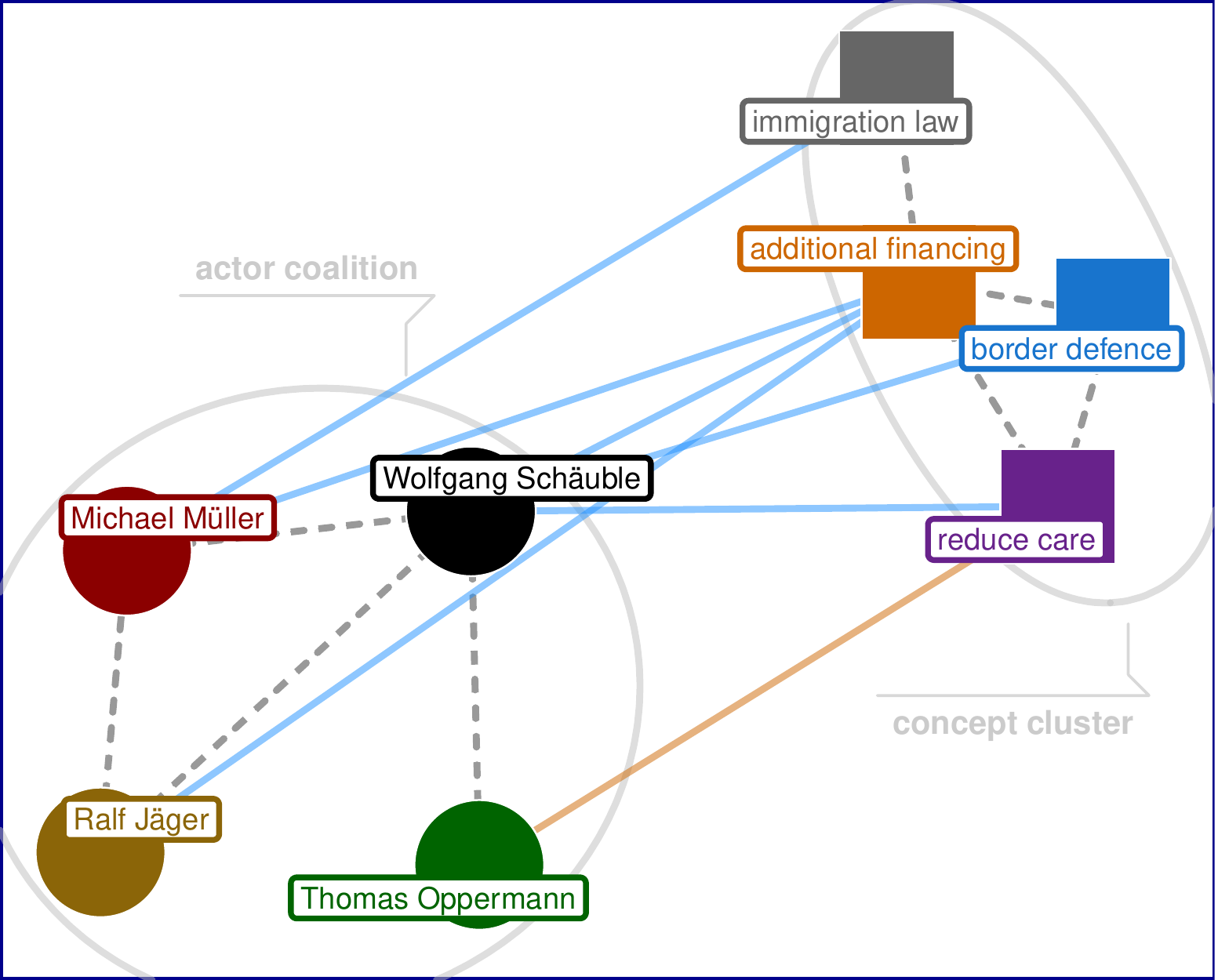}
   \caption{network (4 actors 4 claims); adapted from \citep[p. 71]{janning2009}}
   \label{fig:conceptb}
\end{subfigure}

\caption[]{The annotation process: a) Panel 1 displays four text passages containing political Claims. Claims and actors are highlighted and coloured. Panel 2 depicts the resulting annotations actors (circles), claims (squares) and polarity (plus/minus signs). As shown in b) and c) political claims may contain more than one actor or claim but no diverging polarity. In Panel 3 the information encoded within the annotations are pivoted into (un-stacked) observations with each row encapsulating an actor-claim-polarity triplet. In b) this information is transformed into an (bipartite) affiliation network with claims and actors as two types of vertices in the network. Edges are coloured according to their weight. Dashed and grey lines represent the projected variants of the then one-mode networks.}
\label{fig:concept}
\end{figure}

As mentioned in the introduction, the unit of observation consists of textual entities, so called political claims, and actors (politicians, parties, social movements, NGOs, intellectuals, etc.). In this section, we discuss the conceptual and practical issues related to our annotation, spelling out its different steps.  Recall that the components of a discourse network are nodes (claims and actors) and edges. Our annotation takes raw text as an input, and aims at identifying these components and characterize them in a multi-layered fashion.  The  annotation process is visualized in Figure \ref{fig:concept}, in panel (1) and (2). After identifying relevant text spans (1), annotators then attribute both the corresponding actor(s) as well as choose claim categories, polarities and date (2). Note that the annotation steps discussed in this section are not specific to a DNA approach – the additional operations necessary to get to a DNA are illustrated in section \ref{subs-postproc}. The core annotation steps are: claim identification (step 1) and classification (step 2); date assignment (step 3); actor identification (step 4); claim attribution (step 5) and polarity assignment (step 6).   

\noindent
\textbf{Step 1: claim identification.} First, the annotators have to identify the textual spans containing claims. As shown in our examples in Figure 1, claim-bearing textual spans do not necessarily coincide with a sentence: they can be a subpart of a sentence, or span beyond the sentence boundary. Discourse network-wise, this corresponds to the identification of the claim node (squares in Figure 1). The claim node is further annotated with the document ID. 

\medskip
\noindent
\textbf{Step 2: claim classification.} The textual spans identified above are assigned to one or more claim categories from the annotation scheme (called \textit{codebook} in Political Science), which structures the annotation process. Categories and annotations are moving targets, they traverse a `hermeneutic cycle' during the annotation process: the aptitude of individual claim-categories is constantly reviewed, new categories are adopted, outdated categories revised, overlapping categories merged. All this requires a close cooperation between experts and trained annotators in order to maximize inter-coder reliability.

\begin{table}[t]
\small
\caption{\label{tab:cats} High-level categories across annotations}
\centering
\begin{tabular}[t]{rlrr}
\toprule
major & description & frequency & percentage\\
\midrule
100 & Controlling Migration & 992 & 22\\
200 & Residency & 630 & 14\\
300 & Integration & 386 & 9\\
400 & Domestic security & 154 & 3\\
500 & Foreign Policy & 711 & 16\\
600 & Economy + Labour Market & 153 & 3\\
700 & Society & 740 & 17\\
800 & Procedures & 651 & 15\\
\bottomrule
\end{tabular}
\end{table}

Overall, 110 detailed sub-categories exist that fall into 8 high-level classes shown in Table \ref{tab:cats}. Besides fine-grained categories, the codebook also contains descriptions and defining examples providing guidance to our annotators (an example is provided in table \ref{tab:categories}, appendix \ref{appendixB}); the codebook can be downloaded from \url{http://hdl.handle.net/11022/1007-0000-0007-DB07-B}). Claims are unevenly distributed across classes paying tribute to foci of the discussion at hand.

The claim nodes identified at the previous step are now characterized by two annotation layers, namely the fine-grained and the coarse-grained categories. 

\medskip
\noindent
\textbf{Step 3: date assignment.}  the claim is assigned a date, which is by default the day preceding the publication of the article. It is the annotator's task to reconstruct the claim date, based on textual information. The assigned date is a further layer in the annotation of the claim nodes.

\medskip
\noindent
\textbf{Step 4: actor identification and mapping.} The annotators identify the strings corresponding to actor mentions (e.g., `Angela Merkel', `Die Kanzlerin', `Frau Merkel'). Note that a single claim can be attributed to more than one actor, and actors can be mentioned inside or outside the textual span. Discourse network-wise, this step corresponds to the identification of the actor node (circles in Figure 1). The actor nodes are further annotated with: a) named entity (PER vs. ORG),  b) party\footnote{Party affiliations are either directly annotated, if provided within the newspaper article, or attributed ex post (either manually or by querying Wikidata).}, for the politicians, c) mapping of the actor mention to a canonical name which serves as a unique identifier of the actor in the dataset (e.g., `Angela Merkel' for `Die Kanzlerin').

\begin{table}\small
\caption{\label{tab:map}actor mapping}
\centering
\begin{tabular}[t]{llll}
\toprule
 & pattern & example 1 & example 2\\
\midrule
1 & string & A. Merkel & mehrere Ministerpräsidenten\\
2 & spelling & Angelika Merkal & mehrere Ministerprasidenten\\
3 & parts & Merkel & Ministerpräsidenten\\
4 & synonym & Kanzlerin & Länder-Chefs\\
5 & canonical name & Angela Merkel & mehrere Ministerpräsidenten\\
\midrule
6 & surrogate & Spokesperson & Spokepersons\\
7 & pars pro toto & Bundesregierung & \\
8 & totum pro parte &  & alle Bundesländer\\
9 & possible name & Bundesregierung & list of all names\\
\bottomrule
\end{tabular}
\end{table}

Table \ref{tab:map} 
contains two examples highlighting potential difficulties that may arise during the process of actor mapping. While Example 1 refers to the actor by name, Example 2 points to the institutional role/position of the actors instead. 
 The top half of the table (rows 1-5) displays a set of operations which fall under the general label of `lexical mappings'. Broadly speaking, the goal of lexical mapping is to find the canonical name of the actor. This implies unifying different notations, rectifying spelling errors, completing names, and resolving synonyms. At the end of this process, `Angela Merkel' emerges as the canonical name in Example 1 and `multiple prime minister' in Example 2. While the former is a proper actor, the latter obviously still needs to be resolved to individual ministers. But what if we simply cannot know which prime ministers are being referred to here? In this case, we can either stick with the somewhat unsatisfying canonical name that we found or settle on a set of mapping rules. For instance, we could map the prime ministers as heads of their corresponding states to the group-actor `Bundesländer' representing the states in Germany. 
 
 The bottom half of table \ref{tab:map} (rows 6-9) illustrates even more problematic cases, in which mapping requires assume more complex considerations that are less easy to formulate in terms of mapping rules. This is the case of spokespersons or institutional actors that represent another actor. Does the spokesperson of the federal government also speak for Angela Merkel as the chancellor and leader of the government? If so, we blur the lexical with the semantic level. If not, we might lose a lot of claims that are concerned with the daily business of running a country. On-going work aims to generalize mapping rules at different levels of granularity depending at the task at hand. 

 \textit{DebateNet2.0} contains only the former type of mapping, i.e. lexical mapping.

\noindent
\textbf{Steps 5 and 6: claim attribution and polarity assignment.} The claims are now explicitly linked to the relevant actor. Discourse network-wise, this step corresponds to establishing an edge between the claim and actor nodes. The last step is to assign a polarity to the edge: does the actor support or reject the categorized claim? Polarity assignment is thus performed as an edge labelling operation.

\subsection{From texts to discourse networks}  
\label{subs-postproc}
In the previous section, we discussed the annotation process and its multiple layers for actors and claims. A discourse network is not the only way to query the dataset, obviously: plenty of descriptive statistics and qualitative insights can be gained from a purely quantitative analysis, which we discuss in the next section. To get to a discourse network, however, three further steps need to be performed, which we discuss below. 

\noindent
\textbf{Step 7: Unstacking.}  Given that one text-span may consist of multiple categories and include several actors the format usually required for DNA needs to be achieved first by un-stacking the  annotation. In practice, this means that for each multi-class claim node previously identified, duplicated nodes are created which contain an identical span and date, but a single claim category at a time. The procedure is outlined in Figure \ref{fig:concepta}, panel 3.

\begin{figure}[t]
 \centering
    \includegraphics[width=.9\linewidth]{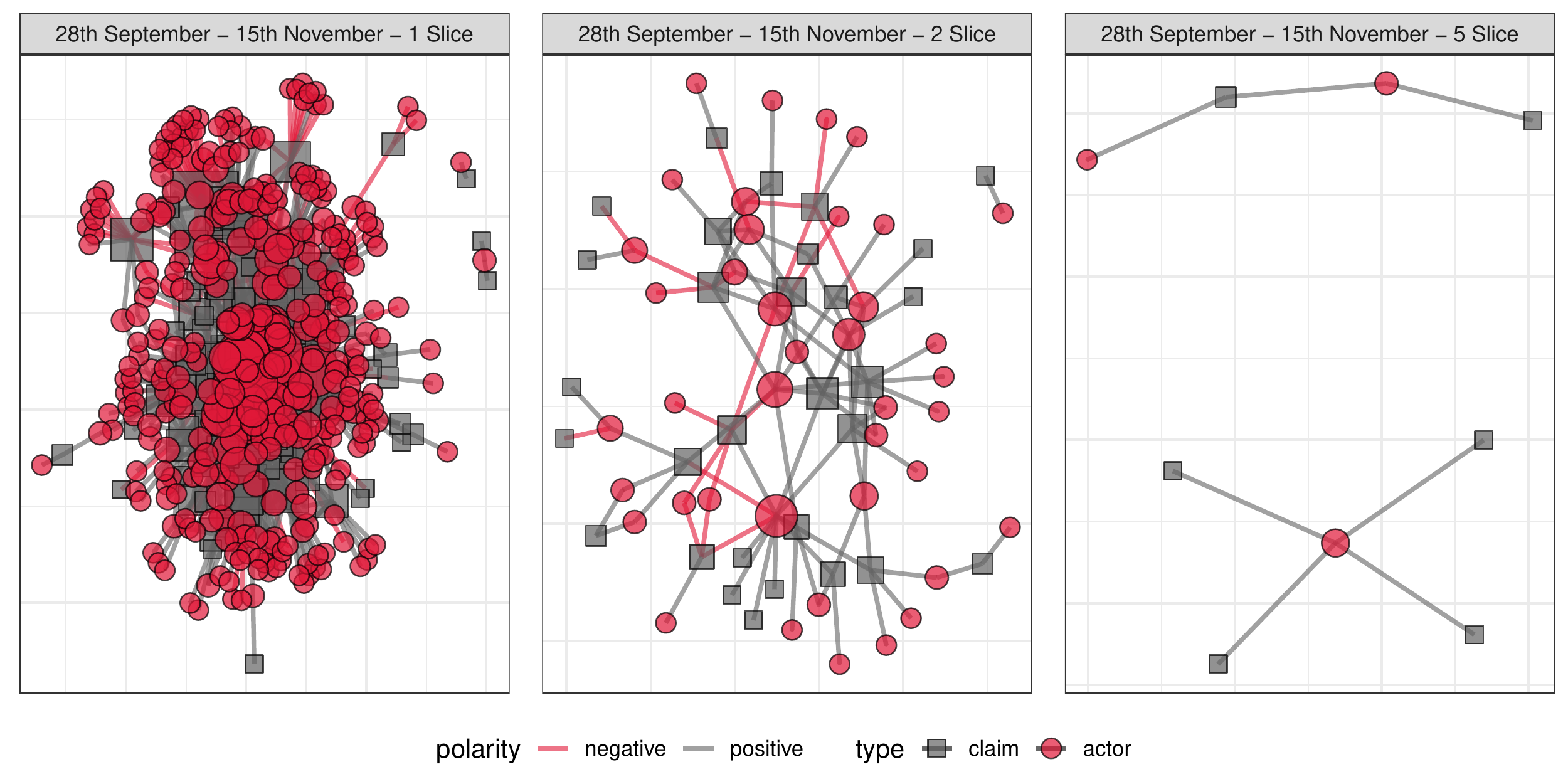}
   \caption{Defining the core network: 1-slice (left), 2-slice (middle), 5-slice (right)}
   \label{fig:slice} 
\end{figure}

\noindent
\textbf{Step 8: Aggregation.} Once the multi-class claims have been unstacked, aggregation can be performed on the resulting representation, following the desired criteria which target the different annotation levels. Aggregation is often performed in multiple steps. The typical first step is to aggregate across time-windows, and perform a first filtering on the actor/claim pairs which occur at least \textit{m} times, a typical value of \textit{m} being 2. This step builds on the assumption that, given the redundancy of the reports, hapaxes at an actor/claim pair can be considered spurious and can be trimmed away without losing relevant information. The resulting network is referred to as the `core' network \citep[p. 112-113]{scott2000}, or, more technically, as a \textit{m-slice} of the complete network \citep[p. 98]{denooy2005}. The intuition behind the core network is to extract only those nodes and edges that are very central to the debate and filter out potential noise. For purely illustrative purposes, we display in Figure \ref{fig:slice} the 1-slice (raw), 2-slice and 5-slice network for the same time period (circles are actors, squares are claims). The 1-slice network is extremely dense, contains a high number of actors and claims that appear at least once, and is therefore often difficult to interpret. In contrast, in the 5-slice network very little of the structure of the debate is left (indeed, it is an extreme case that an actor/claim would be mentioned at least 5 times on different days within the specified time-window). In the following sections, we will adopt a 2-slice representation, which generally strikes an appropriate balance between relevant information and comprehensiveness. 

After the hapax claim/nodes pairs are filtered away on a day-base, typically a further aggregation step comes into play, which targets the node annotation: e.g. aggregation on a monthly base, or on the bases of a claim category, or actor, or a combination of them. 

\noindent
\textbf{Step 9: Projection.} Once the claims have been unstacked and aggregated according to a desired criterion, yet another level of abstraction may be applied to capture conceptual similarity: the actor-claim dyad can be transformed into an actor-actor or claim--claim relation by projecting the claim nodes on the actor nodes, and the other way around \citep{leifeldDiscourseNetworkAnalysis2016}. Figure \ref{fig:conceptb} (page \pageref{fig:conceptb}) illustrates how the projection of the actor nodes on the claim nodes gives rise to the so-called concept-cluster, while the opposite operation leads to the identification of actor coalitions. 

Depending on the research question, distinct follow-up transformations and the introduction of arbitrary temporal boundaries are feasible \citep{leifeldDiscourseNetworkAnalysis2016}. Some possible scenarios are covered in section \ref{DNA}. 
\textit{DebateNet 2.0} is in reality the result of going several times through steps 1 to 9, but could in principle be archived in one single pass. Section \ref{descriptive} provides an overview of some core text statistics of the dataset.

\subsection{Descriptive statistics of \textit{DebateNet 2.0}}
\label{descriptive}

In this section we discuss descriptive statistics for \textit{DebateNet 2.0}, at the level of actors, claims, and textual spans. This section is thus not concerned with Discourse Network Analysis, for which we refer the reader to section \ref{DNA}. 

\noindent The dataset contains 3442 text passages (2. panel Figure \ref{fig:concepta}), which translate into 4417 individual claims (3. panel Figure \ref{fig:concepta}).\footnote{Panel 3 Figure \ref{fig:concepta} actually combines two steps into one: The un-stacking of actors and of claims. Note, that in the provided dataset the un-stacking of the actors was already carried out; therefore, the dataset contains 3861 observations at this interim-step.} 
\paragraph{Actor level statistics.} Table \ref{tab:actors} displays the 10 most frequent actors of the entire year. Unsurprisingly, the most prominent actor of the migration crisis in Germany is chancellor Angela Merkel, followed by minister of the interior Thomas de Maizière, and, as institutional actor, the federal government (`Bundesregierung'). 
Other relevant actors include Merkels' political antagonist in the migration debate Horst Seehofer and chairman of the Christian Democratic Union (CSU), the Foreign Minister Sigmar Gabriel, and also the President of the European Commission Jean-Claude Juncker.

\begin{table}
\small
\caption{\label{tab:actors}Frequent actors in the migration debate 2015 (after actor mapping)}
\centering
\begin{tabular}[t]{lr}
\toprule
\multicolumn{2}{c}{Actors} \\
\cmidrule(l{3pt}r{3pt}){1-2}
name & freq\\
\midrule
Angela Merkel & 247\\
Thomas de Maizière & 162\\
Bundesregierung & 152\\
CSU & 86\\
Horst Seehofer & 79\\
SPD & 78\\
EU & 77\\
Sigmar Gabriel & 68\\
Grüne & 60\\
Jean-Claude Juncker & 57\\
\bottomrule
\end{tabular}
\end{table}

\paragraph{Claim-level statistics.} Recall that Table \ref{tab:cats} (page~\pageref{tab:cats}) describes the distribution of higher-level-categories across the dataset. The most frequent category is `controlling migration', which contains demands and proposals concerned with regulating immigration (border controls, upper limit, asylum law, etc.). Related and also prominent is `foreign policy' (e.g. EU-wide quota, international solutions). Other categories are `society' that deals with humanitarian and cultural aspects (human rights, Christian values) and `residency', mostly concerned with the accommodation of migrants. Least frequent are `domestic security' and `economy + labour market' which are further downstream of the acute (perceived) crisis situation. Although, it stands to reason that both gained momentum after 2015 and are more present during the 2016 discussion (not covered in our dataset). A special, less topical category is `procedures' that often appears in combination with other categories (additional funding, transparency, etc.). 
Table \ref{tab:claims_pos} and \ref{tab:claims_neg} 
display the most frequent positive and negative sub-categories of 2015, respectively. EU-Solution (501), such as a European-wide quota for refugees, is the most often used claim with a positive polarity. Followed by calls for more funding (805) and an upper limit (102). Oppositely, frequent claim categories with negative polarity are calls to oppose xenophobia (703), right wing radicalism (709), and the current immigration policies (190). An EU solution is also high on the list, indicating that this is a polarizing and contested claim category.

\begin{table}
\small
\caption{\label{tab:claims_pos}Frequent Positive Claims}
\centering
\begin{tabular}[t]{llrr}
\toprule
\multicolumn{4}{c}{Positive Claims} \\
\cmidrule(l{3pt}r{3pt}){1-4}
Code & Claim Category & Frequency & Total (incl. neg. Claims)\\
\midrule
501 & EU solution (quotas for refugees) & 237 & 310\\
805 & additional financing & 147 & 155\\
102 & ceiling/upper limit & 129 & 152\\
812 & fast / accelerated procedure & 124 & 132\\
207 & deportations & 112 & 140\\
504 & safe country of origin & 112 & 153\\
105 & border controls & 103 & 126\\
705 & refugees welcome & 94 & 107\\
309 & care (medical, financial, ...) & 87 & 119\\
104 & isolation/immigration stop & 86 & 128\\
\bottomrule
\end{tabular}
\end{table}

\begin{table}
\small
\caption{\label{tab:claims_neg}Frequent Negative Claims}
\centering
\begin{tabular}[t]{llrr}
\toprule
\multicolumn{4}{c}{Negative Claims} \\
\cmidrule(l{3pt}r{3pt}){1-4}
Code & Claim Category & Frequency & Total (incl. pos. Claims)\\
\midrule
703 & xenophobia & 129 & 161\\
709 & right-wing radicalism & 86 & 98\\
190 & current migration policy & 73 & 95\\
501 & EU solution (quotas for refugees) & 73 & 310\\
202 & refugee accommodation & 56 & 79\\
110 & asylum law & 49 & 98\\
711 & islam & 47 & 63\\
104 & isolation/immigration stop & 42 & 128\\
504 & safe country of origin & 41 & 153\\
401 & violence against migrants & 36 & 44\\
\bottomrule
\end{tabular}
\end{table}

\begin{figure}[t]
 \centering
   \includegraphics[width=0.9\linewidth]{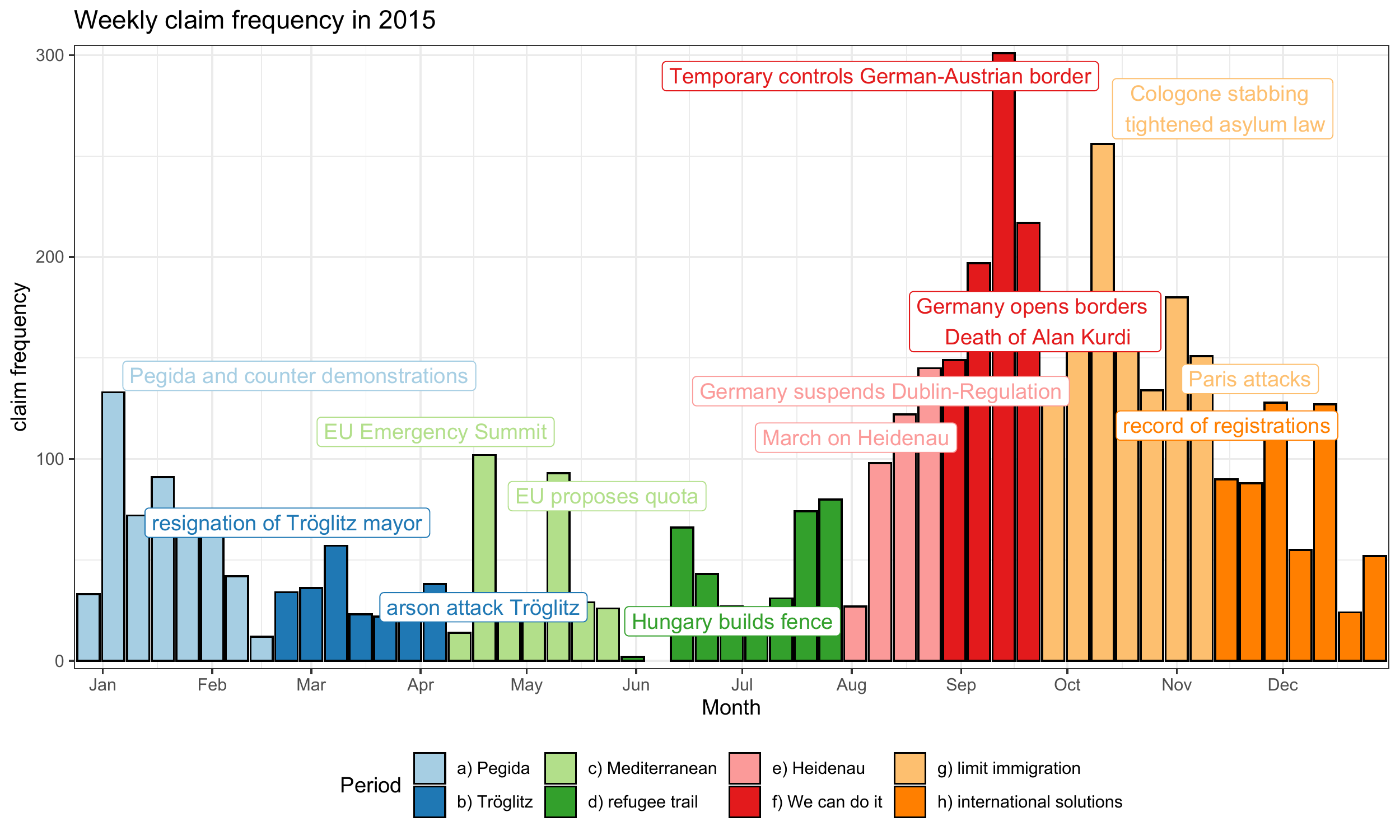}
   \caption{Timeline of the `refugee crisis' in Germany. }
   \label{fig:timeline_detailed}
\end{figure}
\paragraph{Timeline.} Before moving to DNA, we showcase the number of resulting claims over the course of 2015 in Figure \ref{fig:timeline_detailed}. The graph displays the weekly number of claims in 2015 roughly divisioned into time periods named after prominent events (modified after \citealt{haller2017})\footnote{Haller identifies 10 pertinent events and investigates how they were conveyed in leading German newspapers. Here, we adopted some parts and labels of the detailed classification of this study and transferred them to a much coarser categorization. For a brief summary of each time period refer to appendix \ref{appendixA} (page~\pageref{appendixA}).}\footnote{Labeled dates symbolizing micro-events are constructed following \url{https://en.wikipedia.org/wiki/Timeline_of_the_European_migrant_crisis#2015} and \url{https://www.sueddeutsche.de/politik/migration-etappen-der-fluechtlingskrise-in-deutschland-eine-chronologie-dpa.urn-newsml-dpa-com-20090101-160831-99-279450}}. This is not meant to be a full-fledged analysis, but a rough classification that guides future examinations. 
As foreshadowed by Figure \ref{fig:timeline} (page~\pageref{fig:timeline}), claims are unevenly distributed throughout the year, with distinct peaks in September and October. 
Each phase transition is a natural way to separate the debate into fragments that in turn can be analyzed in depth separately or in combination with other periods. For instance, with the numbers of refugees trying to escape the war in Syria via the Mediterranean the discussion about the necessity of sea rescues and efforts to fight smugglers flare up in April. A debate less critical by the time the `Balkan Route' gains prominence. Obviously, there is causal connection between relocating the points of entry into the European Union but the focus of the discussion has shifted to other policies to address the new developments (e.g. the Hungarian border fence `Grenzzaun').

\paragraph{Text-level statistics: keyword analysis.} An alternative way to explore and analyse the data provided in \textit{DebateNet2.0} is using keywords. This approach builds on the assumption that changes in policies are accompanied by a different vocabulary to express them. In other words, are these changes reflected in a change of word distributions?
To extract the keywords, we used the YAKE system \citep{Campos2018} with standard settings. YAKE's keyword extraction approach combines word frequency information with finer-grained features related to the position of a candidate keyword in the sentences in which it occurs, as well as with dispersion scores (at the sentence level and at the level of the different contexts of occurrence). For each annotated text passage, we extracted two keywords with YAKE. We then aggregate the keywords per month, ranking them according to  frequency.
Table \ref{tab:keywords} (page \pageref{tab:keywords}) displays the most frequent keywords for the month April (representing the \textit{Mediterranean} period) and October (representing the \textit{Balkan route} period). As expected, the word `Seenotrettung' (emergency sea rescue) and `Schlepper' (smugglers) are most frequent in April along with `Bundesregierung' (federal government). The latter also  occurs frequently in October, together with all governmental parties (CDU/CSU and SPD) and high ranking government officials. As for claims, keywords indicating restrictionist policies, such as `grenze/grenzen' (border), `transitzonen' (transit zones), and `abschottung' (isolation/walls-up) rank high in October.

\begin{table}[t]
\small
\caption{\label{tab:keywords} Most frequent keywords in April and October.}
\centering
\begin{tabular}{lrlr}
\toprule
\multicolumn{2}{c}{April} & \multicolumn{2}{c}{October} \\
\cmidrule(l{3pt}r{3pt}){1-2} \cmidrule(l{3pt}r{3pt}){3-4}
keyword & freq & keyword & freq\\
\midrule
bundesregierung (fed. government) & 8 & spd (SPD) & 20\\
seenotrettung (sea rescue) & 6 & flüchtlinge (refugees) & 19\\
cdu (CDU) & 5 & cdu (CDU) & 18\\
maizière (Maizière) & 5 & merkel (Merkel) & 17\\
europa (Europe) & 4 & deutschland (Germany) & 13\\
schlepper (smugglers) & 4 & csu (CSU) & 12\\
spd (SPD) & 4 & bundesregierung (fed. government) & 11\\
asylbewerbern (asylum seekers) & 3 & regierung (government) & 10\\
aktivisten (activists) & 2 & flüchtlingen (refugees) & 9\\
asyl (asylum) & 2 & seehofer (Seehofer) & 8\\
behandlung (treatment) & 2 & transitzonen (transit zones) & 8\\
bitte (request) & 2 & grenze (Border) & 7\\
bleiberecht (Right of residence) & 2 & oktober (October) & 7\\
bundesamt (federal office) & 2 & abschottung (isolation) & 6\\
bundesländern (states) & 2 & grenzen (borders) & 6\\
bündnis (alliance) & 2 & kanzlerin (chancellor) & 6\\
fähren (ferries) & 2 & land (country) & 6\\
flüchtlinge (refugees) & 2 & maizière (Maizière) & 6\\
rettungsmission (rescue mission) & 2 & türkei (Turkey) & 6\\
\bottomrule
\multicolumn{4}{l}{\textsuperscript{a} For October we present all keywords with a frequency above 5.}\\
\multicolumn{4}{l}{\textsuperscript{b} In April this threshold is 2.}\\
\multicolumn{4}{l}{\textsuperscript{ } Note, that we hand-picked the keywords in case there was a tie in frequencies.}\\
\end{tabular}
\end{table}

These findings demonstrate that the keyword lists can provide a general idea about the topics of discussion and the most central actors involved. Yet, it is not possible to make inferences regarding the relation between different actors and claims on this basis. This motivates our use, in the next sections, of discourse networks. We expect that they not only  mirror these developments, but also shed some light onto relational aspects and the decision making along the process with it. 

\begin{figure}[t]
\centering
   \includegraphics[width=.75\linewidth]{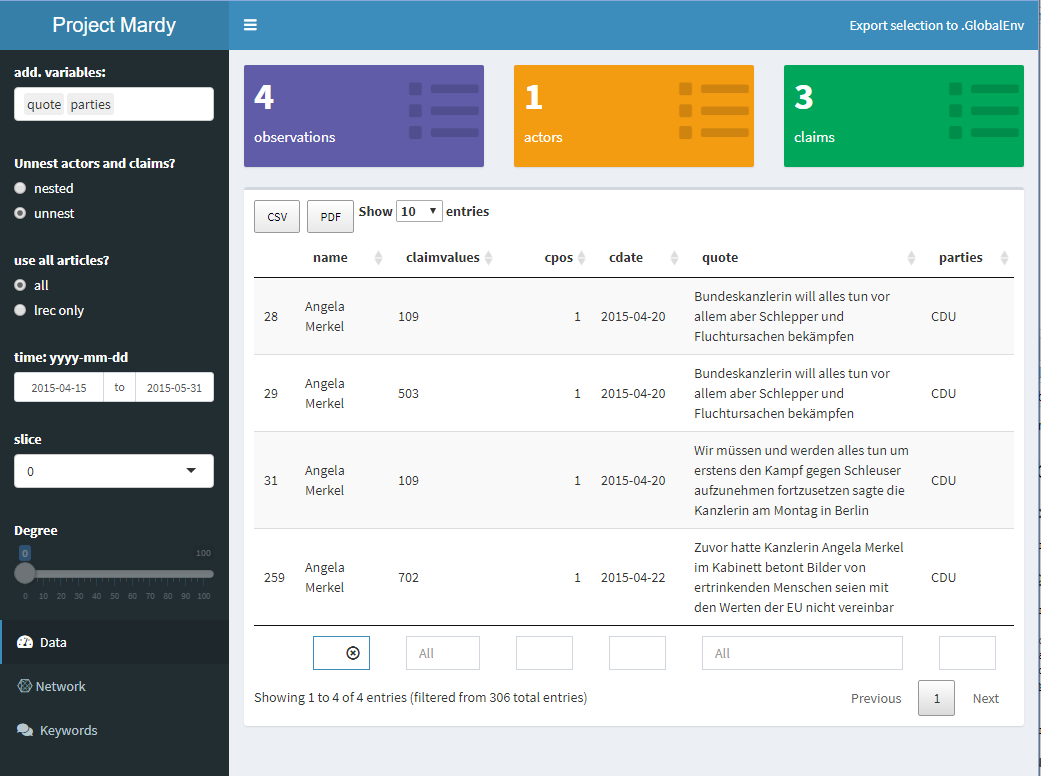}
   \caption{mardyR - dashboard}
   \label{fig:dashboard} 
\end{figure}

\section{Visualization with \texttt{mardyR}}
\label{mardyr}

In order to quickly inspect the dataset and replicate the post-processing steps described in Section 3, as well as to obtain a visualization of discourse networks (Section 5), we developed a companion software package in \texttt{R}, \texttt{mardyR}. The package offers an interactive, browser-based web application in the form of a dashboard built around the shiny\footnote{https://shiny.rstudio.com/ and https://rstudio.github.io/shinydashboard/} framework \citep{shiny,shinydash}\footnote{The app and a brief manual can be found here: \url{http://hdl.handle.net/11022/1007-0000-0007-DB07-B}}.

\begin{figure}[t]
\centering
   \includegraphics[width=.55\linewidth]{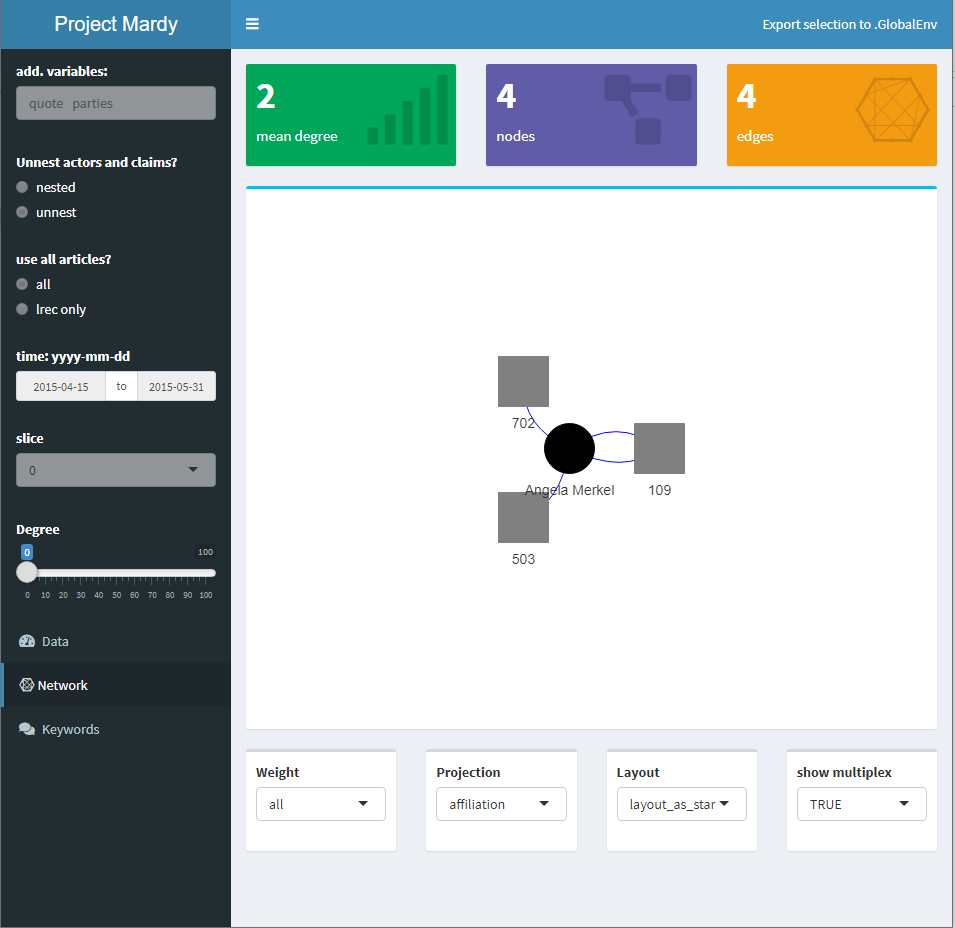}
   \caption{Spring: Ego-network Merkel}
   \label{fig:merkel1} 
\end{figure}

\textbf{Actor-level aggregation.} For demonstration purposes, we use this application to trace the involvement of chancellor Angela Merkel over the course of the debate. We selected the time-windows from Figure \ref{fig:timeline_detailed}, corresponding to `Mediterranean' and `Limit Immigration'). Figure \ref{fig:dashboard} (page \pageref{fig:dashboard}) displays all (un-stacked) claims by Merkel as reported in the \textit{taz} from 15th April to 31st May. The entries of the dataset are presented in the right panel of Figure \ref{fig:dashboard} while the left panel contains configuration options, e.g. select variables, apply time-frame, choose which release of the dataset to use.  In this period, she stresses the importance of humanitarian rights (702) and proposes to fight smugglers (twice, 109) as well to combat causes of flight (503). The corresponding network is displayed in Figure \ref{fig:merkel1} (page \pageref{fig:merkel1}), with Merkel as round node surrounded by and linked to three square nodes representing claim-categories. Overall, the resulting network consists of four nodes and four edges with an average degree centrality of 2  (measured as the average number of edges connected to a node, \citealt[p. 100]{wasserman1994}).

\begin{figure}[t]
\centering
   \includegraphics[width=.55\linewidth]{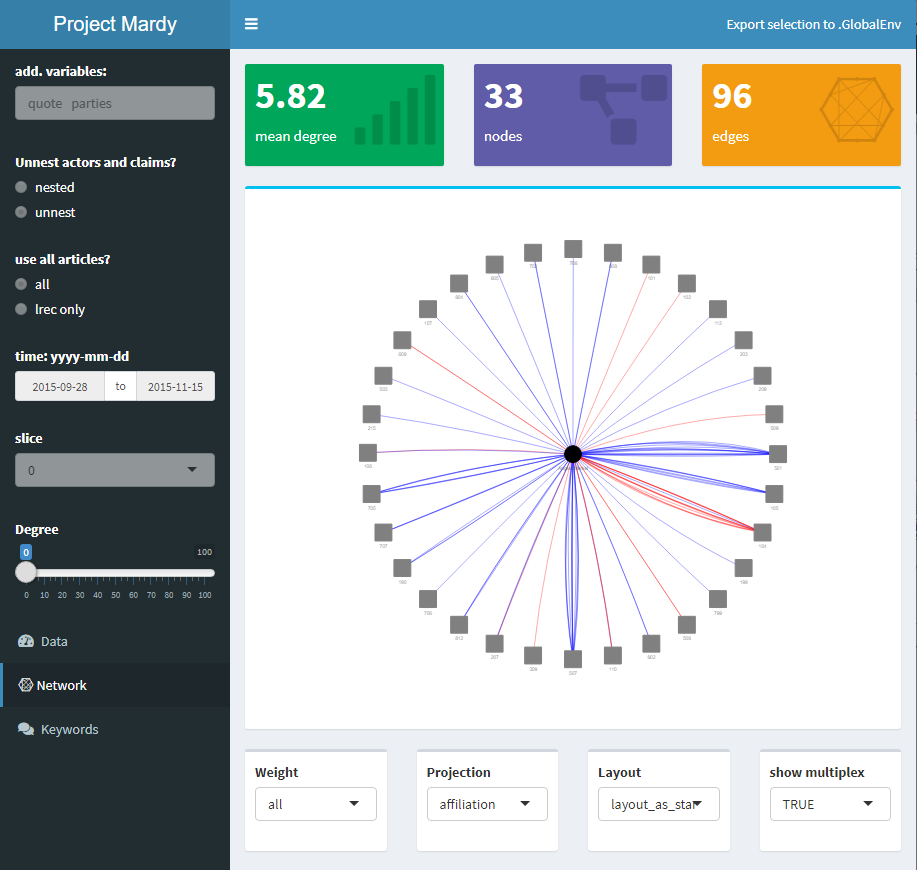}
   \caption{Fall: Ego-Network Merkel}
   \label{fig:merkel2} 
\end{figure}

In contrast, Figure \ref{fig:merkel2} (page \pageref{fig:merkel2}) makes it apparent how much more involved she became during the later period of the year from 28th September to 15th November. Suddenly, 96 claims made by Merkel are reported in the newspaper, which translate into 32 distinct claim categories she either supports (blue edges) or rejects (red edges).

\begin{figure}[t]
\centering
   \includegraphics[width=.55\linewidth]{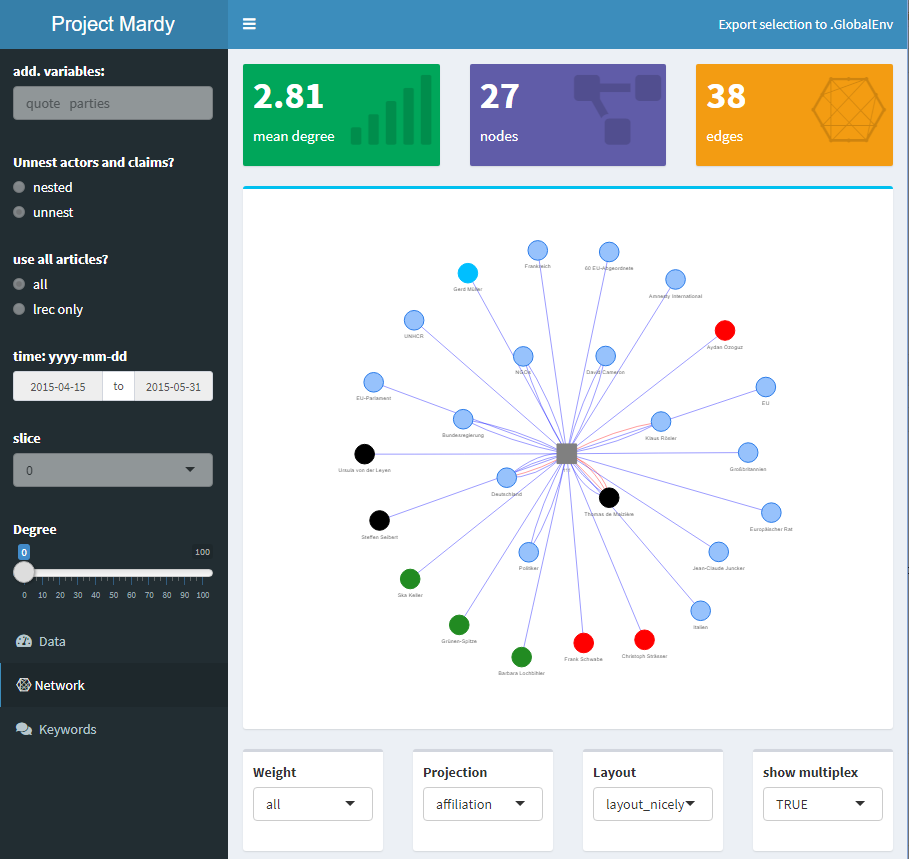}
   \caption{Spring: Ego network sea rescue}
   \label{fig:111} 
\end{figure}

\textbf{Claim-level aggregation}: Similarly, one can trace the usage of different claim categories over time or cross-sectionally. In Figure \ref{fig:111} (page \pageref{fig:111}) all instances of the claim `sea rescue' (111) from the first time period are displayed and linked to the corresponding actors (colours indicate party affiliation). The overwhelming majority supports the claim of sea rescue missions. Therefore, from a purely structural standpoint, it is even more surprising that this claim does not appear in the second time period at all. Instead of 111 other claims have emerged and gained popularity. 

Overall, this goes to show the dyadic relation between actors and claims and vice versa. However, the real strength of social networks only comes to light when the whole network with all its interrelations is considered simultaneously. Only then it is possible to capture the essence of the debate.\footnote{The current version does not support external datasets. However, we plan to update a future version with this functionality. Prerequisite for this is a similar data format (actor - claim - polarity).}

\section{Mediterranean vs. Balkan Route: a Discourse Network Analysis}
\label{DNA}

From the combination of claims of different political actors a network structure emerges, whose analysis brings its own challenges and chances. On the one hand, the analysis becomes more demanding due to the increased complexity of the task, imposed by inherently existing inter-dependencies (actors may influence each other in their decision to oppose or support claims). On the other hand, complexity enables a new relational perspective on the discussion. For instance, it permits the researcher to trace how coalitions and alliances are formed or which actors reach out and bridge between hardened fronts.

We exemplify the use case of discourse network analysis by contrasting two distinct time periods of the 2015 debate in this section: the `Mediterranean' versus the later stages of the `Balkan route' (Figure \ref{fig:timeline_detailed} on page \pageref{fig:timeline_detailed}, `limiting immigration'). Firstly, this allows us not only to assess who the driving actors and claims of the discussion at certain time-points are, but, more importantly, also how they relate to each other and form (dynamic) coalitions. Secondly, we compare the networks obtained by using a) the entire dataset of 2015 (\textit{DebateNet2.0}) and b) a previously released subset of the data (\textit{DebateNet1.0}), in order to evaluate how important the size of the dataset is to understand the migration debate at hand.

\begin{table}\small
\caption{\label{tab:properties}DebateNet2.0 over time}
\centering
\begin{tabular}[t]{lrrrr}
\toprule
month & \# claims & \# unique categories & \# unique actors & average degree\\
\midrule
January & 403 & 61 & 160 & 2.78\\
February & 163 & 46 & 79 & 2.14\\
March & 160 & 52 & 75 & 2.13\\
April & 177 & 47 & 76 & 2.37\\
May & 179 & 43 & 65 & 2.37\\
June & 115 & 44 & 59 & 1.92\\
July & 223 & 54 & 95 & 2.34\\
August & 449 & 64 & 175 & 2.90\\
September & 910 & 90 & 247 & 3.71\\
October & 743 & 78 & 239 & 3.53\\
November & 565 & 80 & 164 & 3.23\\
December & 330 & 69 & 109 & 2.76\\
\bottomrule
\end{tabular}
\end{table}

\paragraph{Network statistics.} A first step in the analysis is to 
 describe the network properties on a basic level. Table \ref{tab:properties}
depicts the number of claim frequencies (second column), the number of unique claim categories (third column), the number of unique actors (fourth column), and the average degree on a monthly basis (last column).
 
 Unsurprisingly, the fluctuating number of claims matches our previous observations (Figure \ref{fig:timeline}, page \pageref{fig:timeline} and Figure \ref{fig:timeline_detailed}, page \pageref{fig:timeline_detailed}).
While September is the month with the most observations, actors and claims, as well as the highest average degree, June operates on a much lower level. Therefore, the intensity of the debate plummets in June, only to increase drastically from August to September. Afterwards it plateaus on a high level until the end of November. 

In the following, we once more focus on the time period discussing sea rescue missions in the Mediterranean from 13th April to 31st of May and contrast it with the debate about limiting migration from 28th September to 15th November in conjunction with the Balkan route,
which ushers in more restrictionist migration policies compared to the preceding `Wir schaffen das' (we can do it) period (see Figure \ref{fig:timeline_detailed}, page \pageref{fig:timeline_detailed}). Additionally, we compare the resulting networks to their counterparts built from the dataset of the previous release with much less observations (\textit{DebateNet1.0}). Recall that this resembles a random sample and its distribution closely follows the current one. 

\begin{figure}[t]
   \centering
   \includegraphics[width=.85\linewidth]{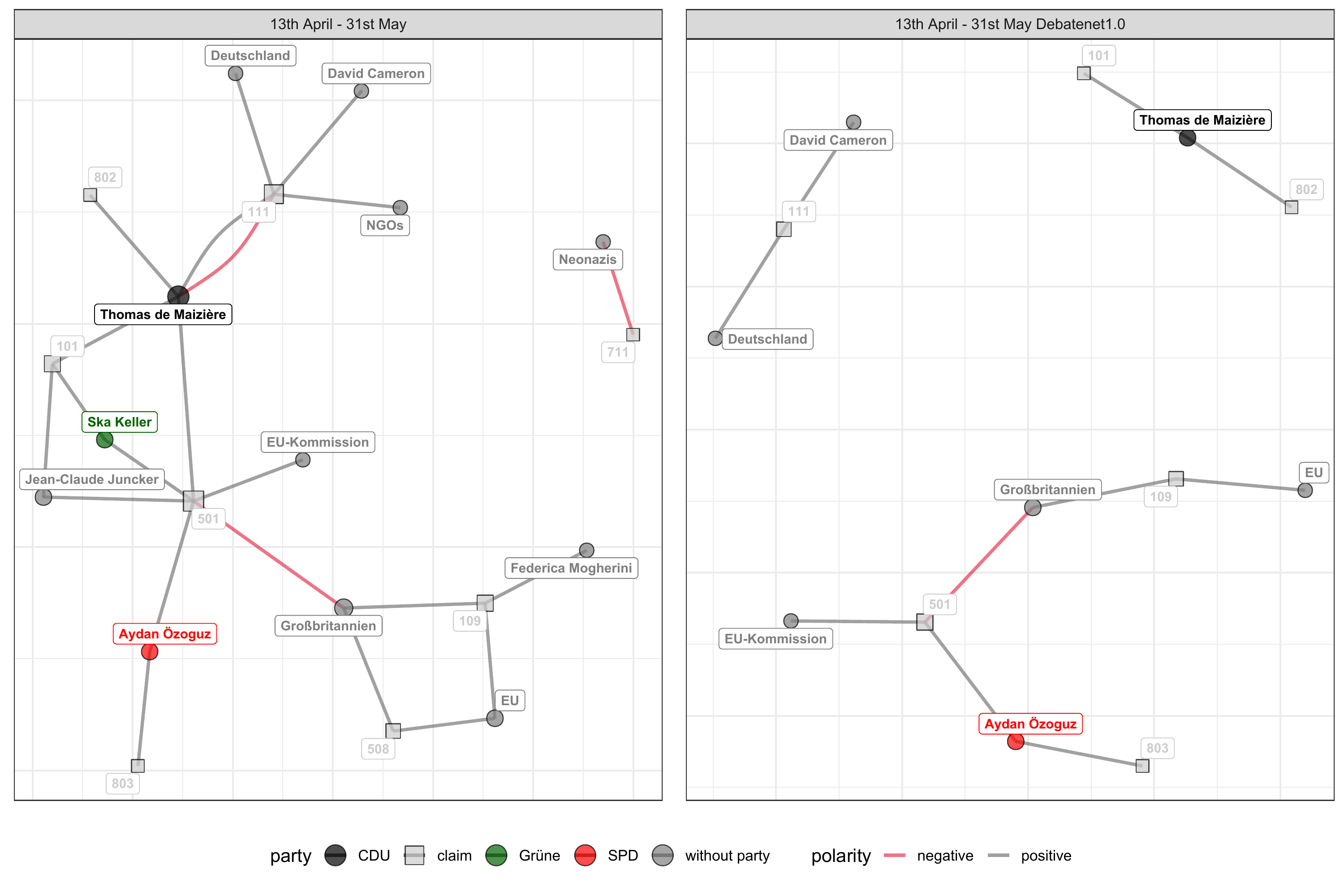}
  \caption{2-slice networks of DebateNet2.0 (left) and DebateNet1.0 (right) from mid April - May}
   \label{fig:janaug} 
\end{figure}

\paragraph{In the spring,} the core network represents a discussion largely at the EU-level (Figure \ref{fig:janaug}, page \pageref{fig:janaug}): The demand to introduce an EU-wide solution (501), possibly a quota for the distribution of refugees, is not only at a very central position in terms of degree but also exhibits the highest betweenness-centrality (measure of how many shortest paths are passing through this node, \citealt{freeman1978a}). Here, it has a brokerage position, bridging two otherwise disconnected components of the network. Similarly prominent is Thomas de Maizière, the minister of the interior, who has a conflicted position with respect to the claim of sea rescue missions (as indicated by the grey and red arc) but supports a EU-wide solution. As does Jean-Claude Juncker and Ska Keller (MP of the Greens). Another relevant claim is that of fighting against people smugglers (109) that finds support from Great Britain (Großbritannien), the EU, and Federica Mogherini (High Representative of the Union for Foreign Affairs and Security Policy of the EU). In the scope of the discussion about fighting against smugglers it was also suggested to launch a military operation to destroy their boats on the coast of Libya, hence the closed fourcycle with category 508 (military intervention). The national level plays only a small part in the periphery of the network: neo-fascists (Neonazis) opposing Islam (711) which ties in with the Islamophobic Pegida (Patriotic Europeans Against the Islamisation of the Occident) movement. Overall, these observations concur nicely with our expectations from the keywords-analysis and proposed time-period: Both topic and actors appear to be largely congruent across methods.

If compared to the network built from \textit{DebateNet1.0} (right panel Figure \ref{fig:janaug}),
the vulnerability of networks \citep{KOSSINETS2006} becomes visible: The core network observed for the complete dataset (left panel, Debatenet2.0) gets fragmented into components that are no longer connected to each other.
By looking at the right panel one might assume that the actors appear not to talk about the same policies at all. This is of course a question of aggregation and the time windows used. Yet it highlights the importance of a comprehensive analysis: If the task is to analyse the core network of the debate, then a random sample would not suffice in this case.

\begin{figure}[t]
   \centering
   \includegraphics[width=.85\linewidth]{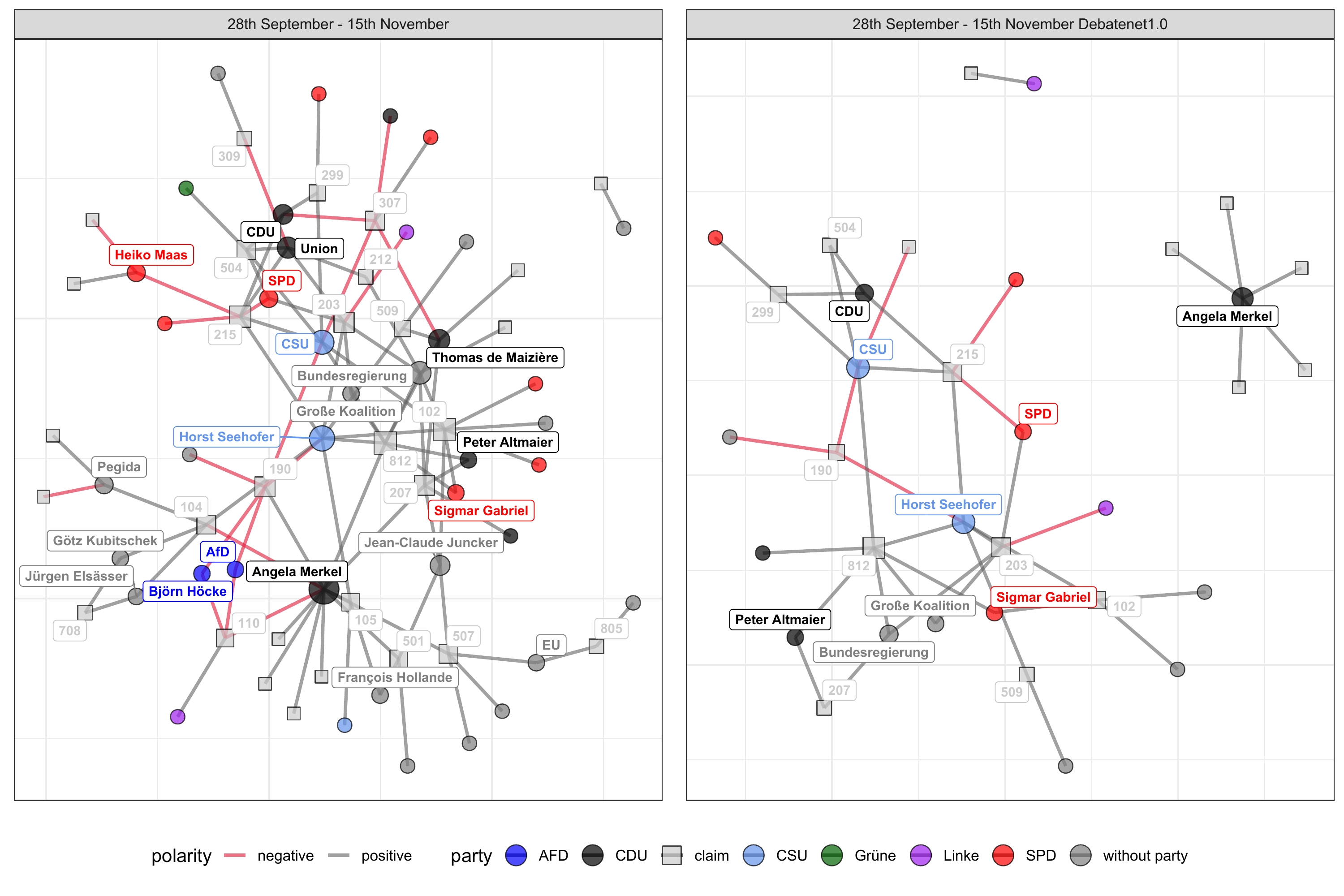}
   \caption{2-slice networks of DebateNet2.0 (left) and DebateNet1.0 (right) from late September - mid November}
   \label{fig:janaug_2} 
\end{figure}

\begin{figure}[t]
\centering
   \includegraphics[width=0.88\linewidth]{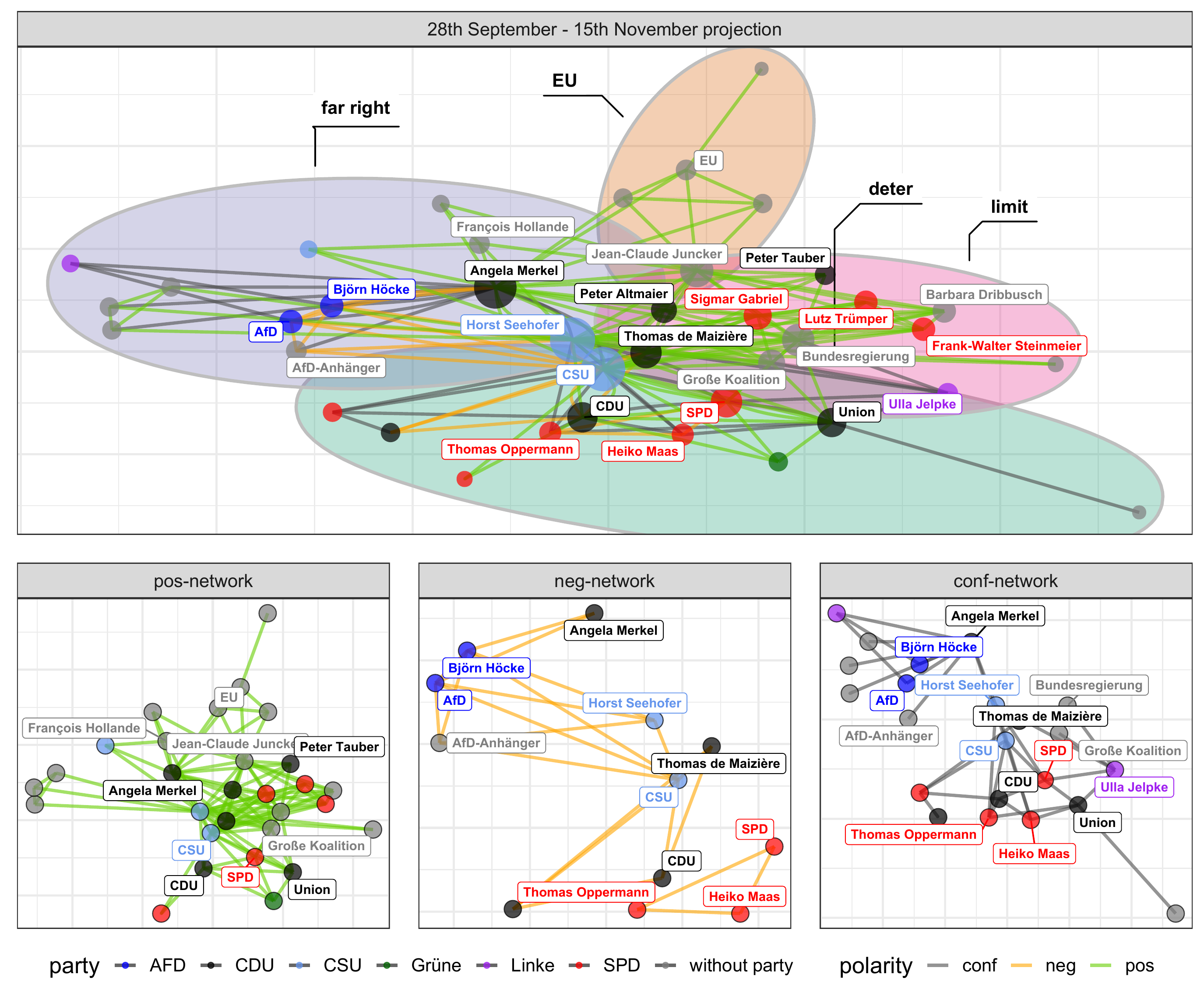}
   \caption{First row: Projection of 2-slice network of fall 2015 (Figure \ref{fig:janaug_2}). Second row: Projections sorted by polarity. Left: Positive congruence network (pos-network), middle: negative congruence network (neg-network), right: conflict network}
   \label{fig:projection} 
\end{figure}

\paragraph{In the fall,} this fact becomes even more apparent (Figure \ref{fig:janaug_2}, page \pageref{fig:janaug_2}). In the left panel, three overlapping clusters within the debate can be observed. The first and largest cluster dedicates itself to questions of residency and integration and involves mostly governmental actors and parties. Arguably, this cluster contains two separate discussions as indicated by the projection onto actor coalitions (Figure \ref{fig:projection}, page \pageref{fig:projection}); namely one focusing on deportations (207) and an upper limit (102) and another discussing so called transit zones (215) and safe countries of origin (504) in which deportations are permitted. While deportations and limits regulate the number of refugees allowed into Germany, transit zones aim to deter applicants by restricting their movement. The second cluster contains the far right movement (e.g. the Alternative for Germany (AfD) and the Pegida organization) and their nationalist claims regarding isolation (104), their expression against the current migration policies (190), and xenophobic and Islamophobic slogans. The third cluster combines actors from the EU and the national level with claims regarding an EU-wide solution (501) via quotas or the readmission-pact with Turkey (507) as well as border controls (105). As foreshadowed by Figure \ref{fig:merkel2}, Merkel has an important bridging function between the different clusters by being connected to a lot of different claims from different clusters in this time period. 

However, once we only consider data from the DebateNet1.0, Merkel gets separated from the giant component and the second and third clusters disappear.
This indicates that a random sampling strategy is unsuitable for the, admittedly highly specific, task of extracting a core network of the migration debate. Nevertheless, the sample is by no means unusable. Instead, it highlights the importance of task-dependency. For example, if the goal is to identify a list of relevant claim categories during the time period from 28th September to 15th November, then even the sample might prove sufficient: Almost 80 percent of distinct claim categories existent in \textit{DebateNet2.0} are already present in \textit{DebateNet1.0} (74 out of 95). This number increases to 95 percent if the entire year is considered.

\paragraph{One-mode Projections} of the actors are another way to look at the fall-network that capture co-occurrences within the data \citep{leifeldDiscourseNetworkAnalysis2016}. Actors are linked to each other if they share the same claims and hence form coalitions based on the topical similarity of their political interests (1st row of Figure \ref{fig:projection}). Depending on their polarity with regard to the claims in question, the relation may either reflect mutual support, shared disagreement, or conflicting stances (2nd row of Figure \ref{fig:projection}). 
Conceptually, this represents yet another level of abstraction from two-mode representation and its underlying textual information. Arguably, this makes it easier to identify clusters within the network:

A basic community detection algorithm based on (greedy) modularity optimization \citep{newman2004,Clauset_2004} suggests four co-existing clusters (communities) within the actor-network (coloured ellipsis in the first panel). The aim of the algorithm, belonging to the family of (agglomerative) hierarchical clustering procedures, is to identify groups of nodes with higher density between them than between nodes of different groups, a property called modularity. At the start, each node is placed in its own cluster and then iteratively merged with neighbouring clusters if the conjunction increases modularity.
The clusters are mostly consistent with the observations made above on the two-mode level: Governmental actors (red, black, and light-blue nodes) are mostly distributed into two clusters (limit and deter, pink and green). European actors are centered around yet another cluster (EU, brown). The last cluster is defined by the debate about fundamental questions of migration by far right actors (far right, violet).
The projection approach highlights how dynamic actor coalitions present themselves as the discourse progresses. For instance, Thomas de Maizière, minister of the interior, now focuses more on domestic policies than he did during the sea rescue discussion amidst the EU compound earlier in the year. Moreover, he is densely embedded into the governmental coalition, representing the building of consensus and thus facilitating more decisive actions within the executive. Similarly, Angela Merkel has to defend her earlier `Wir schaffen das' against propositions by the far right AfD that ultimately would become the biggest opposition party in the 2017 election in Germany.

The benefit of the network-based approach becomes particularly evident by looking at the different kinds of relations expressed in Figure \ref{fig:projection}. Without the network perspective, e.g. by looking only at keywords, it would be difficult to infer how different actors relate to each other. This is not to say that keywords do not provide valuable insights on their own. On the contrary, as shown in Table \ref{tab:keywords} (page \pageref{tab:keywords}), relevant actors and claims correspond highly to prominent nodes found in the network (e.g. Thomas de Maizière). In practice, one could initially deploy keyword analysis in order to get a general understanding of the most prominent actors and pressing issues of the debate, followed up by a more detailed DNA approach. 

\section{Conclusion}

In this paper, we have introduced \textit{Debatenet2.0}, a dataset containing expert annotation on the German political debate on the so-called `refugee crisis' in 2015. The core units of our annotations are political claims (e.g., `quotas for refugees should be established') and the actors who made those claims (e.g., `Angela Merkel', `the government', `the protesters'). While \textit{Debatenet2.0} naturally lends itself to several quantitative investigation approaches (e.g., keyword analysis), its primary purpose is to allow for a Discourse Network Analysis of the debate. Indeed, taking the reader on a tour of the basics of DNA, both conceptually and practically, is one goal of this paper. 
Our contributions comprise multiple levels. At the level of resources, we release \textit{Debatenet2.0} and its companion R package \texttt{mardyR}, which allows the user to create independent queries and visualizations. At the methodological level, we provide a step-by-step explanation of the task of deriving discourse networks from textual data, helping the reader to bridge the conceptual gap between the two fields involved in this interdisciplinary research (Political Science and Computational Linguistics). A further contribution is the case study, in which we address the research questions of a) how specific, concrete events in the crisis affected the discourse network representation of the policy debate and b) how data-dependent the discourse network representation is (we compare networks built from the full dataset to those of a random subsample of it). Broadly, we demonstrate how the debate gained momentum by comparing the spring network with the fall network and we highlight the formation of actor coalitions. Moreover, the vulnerability of the core-network in the face of incomplete (sampled) data is addressed. 
Beyond answering the topical research questions, our case study serves also as an illustration of the application of DNA, and can serve as a guideline for the formulation of research questions on \textit{Debatenet2.0} (and comparable datasets) and for the interpretation of the results. 

Beyond its use to address DNA research questions, \textit{Debatenet2.0} also has a clear potential for NLP, which we have only hinted at in the introduction and background section. As a matter of fact, further NLP modeling has been conducted within MARDY, beyond claim detection and classification  \citep{pado19}. This work has targeted the two core ingredients of the network. On the actor nodes, \citep{dayanik20:_maskin_actor_infor_leads_fairer} have improved the robustness of the claim identification by targeting the frequency bias issue: claim identification (and thus any representation built on it) is fair only if low-frequency actors get identified correctly. On the claim nodes, ongoing experiments test different modeling architectures capable of correctly classifying the fine-grained categories encoded in the codebook (recall, approximately 110); note that this research goal is in a sense parallel to the strive to be fair to low-frequency actors, as fine-grained categories often lead to low-frequency issues which in turn affect robustness over time: as shown in our experiments, specific historical events trigger and shape the debate, making a claim infrequent at time $t$ (the training data) extremely relevant at time $t+1$ (on which we want to apply the classifier).

From a Political Science perspective, further work has targeted the evaluation of the discourse network extracted with a NLP claim identifier, in a direct comparison to the network extracted with the expert annotation;  \citet{haunss20:_integ_manual_autom_annot_creat}
 found that, due to the natural redundancy present in newspaper articles, the 2-slice network derived from the claims identified automatically (with a manual filter on false positives) is equivalent to the one developed from fully manual annotation. Besides, they also show that integrating automatic claim identification in the annotation environment would improve inter-annotator agreement, ensuring a more reliable annotation. As far as the robustness of the immigration codebook is concerned, \citet{blokker20:_swimm_tide} successfully expanded the domain of its application to party manifestos, which contain an `officially approved' representation of the position of a party with respect to specific policy issues, and one which lends itself naturally to a comparison to the individual positions of the political actors which are depicted in the newspaper.
 
 Current work on \textit{Debatenet2.0} adds annotation of articles from 2005 and 2010, and annotation of a new topic, namely the pension debate, to the dataset. At the level of annotation layers, we plan to introduce an additional node type, the so-called \textit{frame} node, which encodes the reason provided by the actor to support a claim (e.g., a border fence should not be built because it violates human rights). From a NLP perspective, the introduction of the frame nodes will bring in new challenges closely related to those tackled by the Argument Mining community \citep{ijcai2018-766,eger-etal-2018-cross}, thus strengthening the interdisciplinary potential of the research agenda behind \textit{Debatenet2.0}. 

\bibliographystyle{spr-chicago}       
\bibliography{bib.bib}   


\begin{appendices}

\section{Annotation guidelines: example}
\label{appendixB}

\begin{table}[H]\footnotesize
  \centering
  \footnotesize
  \begin{tabular}{p{0.95\columnwidth}}
\hline
    \textbf{102: Limit immigration/upper limit}
    \begin{itemize}
    \item \textbf{Explanation:} This claim refers both to limiting the number of refugees by means of a fixed maximum number and to unspecified demands for a reduction in the number of refugees.
    \item \textbf{Example:} `There will be no way around a limit and thus an upper limit for immigration.'
    \end{itemize} \\[-1em] \midrule
 \textbf{111: sea rescue}
    \begin{itemize}
    \item \textbf{Explanation:} This claim refers to demands for the rescue of refugees in distress at sea.
    \item \textbf{Example:} `The human rights organisation is calling on the EU to set up a joint sea rescue operation on the Mediterranean see off the Libyan coast and to create significantly more admission places for refugees.'
    \end{itemize}       \\[-1em] \midrule
 \textbf{504: Safe legal status for country of origin}

    \begin{itemize}
    \item \textbf{Explanation:} This claim refers to demands for the extension of the legal status of one or more countries of origin of refugees as `safe'.
    \item \textbf{Example:}  `In addition, the legislator must declare more Balkan countries safe third countries ``to which we can then deport more quickly''.'
    \end{itemize} \\[-1em] \bottomrule
  \end{tabular}
  \caption{Fine-grained
    categories: codebook examples and annotation guidelines}
  \label{tab:categories}
\end{table}

\section{Timeline}
\label{appendixA}

\begin{enumerate}[label = \alph*)]
    \item Pegida (01/01/2015 - 15/02/2015): This time period is characterised by claims made around the Islamophobic Pegida (Patriotic Europeans Against the Islamisation of the Occident) movement that was founded in October of 2014. Prominent claims and points of discussion of this period are Xenophobia, Islam, and Public Debate (should politicians engage with the movement or categorically refuse communication?).  
    \item Tröglitz (16/02/2015 - 12/04/2015): This period revolves around the town of Tröglitz in Germany and the proposal of its mayor to accommodate refugees, inciting (violent) protests and threats by far-right sympathisers and subsequently compelling the mayor to resign.  Prominent claims and points of discussion of this period are refugee accommodation and violence against migrants.
    \item Mediterranean (13/04/2015 - 31/05/2015): During this period, the media increasingly reported on refugees trying to cross the Mediterranean by boat under high casualties. The EU holds an emergency summit and subsequently proposes a quota to distribute the incoming refugees. Prominent claims and points of discussion of this period are EU-solution, emergency sea rescue, and fight against people smugglers.
    \item refugee trail (01/06/2015 - 01/08/2015): While refugees are still trying to reach Europe from the sea, another route increasingly gains the attention of the media: the Balkan route. Hungary starts to build a border fence to fend off asylum seekers. Still, refugees are reaching Germany invalidating the Dublin-III regulation. Prominent claims and points of discussion of this period are (medical and financial) care for refugees, safe countries of origin, and an immigration law.
    \item Heidenau (02/08/2015 - 30/08/2015): As a reaction to a planned refugee shelter, the extreme right and its sympathisers riot and refuse access to the shelter. Emerging claims and points of discussion of this period are right-wing extrism, (accelerated) deportations, and Dublin-regulation.
    \item `We can do it' (31/08/2015 - 27/09/2015): Angela Merkel declares on 31th August `Wir schaffen das', Germany and Austria open their borders (4th September) only to temporary enforce border controls again later that month. Prominent claims and points of discussion of this period are EU-solution, refugees welcome, and border controls. 
    \item limit immigration (28/09/2015 - 15/11/2015): The `refugees welcome' enthusiasm is followed by a tightening of asylum restrictions and an increasingly tense climate. Prominent claims and points of discussion of this period are upper-limit, walls-up policy, borders controls, and the cooperation with transit countries. 
    \item international solutions (16/11/2015 - 31/12/2015): This period largely continues the policy of the previous phase with an slight increase in claims regarding foreign policies (the G20-summit took place in Antalya on 15/16. November). Prominent claims are EU-Solution, cooperation with transit countries, and combat causes of flight.
\end{enumerate}
For a more in-depth analysis of the events and the accompanying media-echo refer to \citep{haller2017}.

\end{appendices}
\end{document}